# Interval Influence Diagrams

**K. W. Fertig** and **J. S. Breese**
Rockwell International Science Center
Palo Alto Laboratory
444 High Street
Palo Alto, CA 94301

**Abstract**

We describe a mechanism for performing probabilistic reasoning in influence diagrams using interval rather than point valued probabilities. We derive the procedures for node removal (corresponding to conditional expectation) and arc reversal (corresponding to Bayesian conditioning) in influence diagrams where lower bounds on probabilities are stored at each node. The resulting bounds for the transformed diagram are shown to be optimal within the class of constraints on probability distributions which can be expressed exclusively as lower bounds on the component probabilities of the diagram. Sequences of these operations can be performed to answer probabilistic queries with indeterminacies in the input and for performing sensitivity analysis on an influence diagram. The storage requirements and computational complexity of this approach are comparable to those for point-valued probabilistic inference mechanisms, making the approach attactive for performing sensitivity analysis and where probability information is not available. Limited empirical data on an implementation of the methodology is provided.

## 1 Introduction

One of the most difficult tasks in constructing an influence diagram is development of conditional and marginal probabilities for each node in the network. In some instances probability information may not be readily available, and a reasoner wishes to determine what conclusions can be drawn with partial information on probabilities. In others cases, one may wish to assess the robustness of various conclusions to imprecision in the input.

The subject of probability bounds has been a topic of interest for a number of years in artificial intelligence. Early users of Dempster-Shafer formalisms were originally motivated by the ability to specify bounds on probabilities [4,6]. Inequality bounds have also been examined by those attempting to bridge between Dempster-Shafer theory and traditional probability theory [5,8,9]. A number of other researchers have attempted to deal with bounds on probabilities within a traditional Bayesian framework [2,18,12,14].

In this paper we develop and demonstrate a means of incorporating imprecision in probability values by specifying lower bounds on input probabilities and using influence diagrams as a means of expressing conditional independence. A number of authors have developed systems which derive probabilistic conclusions, given general *linear* constraints on the inputs [18,20]. These systems typically use linear programming methods repeatedly to propagate constraints through a set of probabilistic calculations. The characterization of constraints as lower bounds allows us to derive a relatively efficient procedure for probabilistic inference, based on successive transformations to the diagram, at the cost of some expressiveness. The implications of these transformations in terms of the sets of probability distributions admitted by the bounds are analyzed in detail.

## 2 Probabilistic Inference with Bounds on Probabilities

In standard probability theory, when $x$ and $y$ are random variables and completely specified probability distributions of the form $p(x|y)$ and $p(y)$ are available, one can calculate precisely the following quantities:

$$p(x) = \int_y p(x|y)p(y)$$

$$p(y|x) = \frac{p(x|y)p(y)}{\int_y p(x|y)p(y)}$$

using the standard formula for conditional marginalization and Bayes rule.[1] In this paper we examine the case where precise probability distributions are replaced by lower bounds. In general constraints on probability distributions can be considered subsets of the space, $P$, of all probability distributions. In [3], we use this structure as the basis of the interval probability theory presented here. However, for the sake of simplicity, we will start with a special class of constraints for lower bounds.

**Definition 1 (Lower Bound Constraint Function)** *The function $b(x)$ is a lower bound constraint function if and only if*

$$\forall x, b(x) \geq 0, \int_x b(x) \leq 1.$$

**Definition 2 (Constraint Set)** *Let $P$ be the set of all probability distributions $p(x)$ on a space $X$. The set $C \subseteq P$ is the **constraint set** associated with a lower bound constraint function $b(x)$ if and only if*

$$C = \{p | p \in P, p(x) \geq b(x) \forall x\}.$$

Expressing constraints in terms of lower bound constraint functions for all values of $x$ allows us to derive an upper bound for the probabilities for discrete random variables.

**Theorem 1 (Upper Bounds)** *If $x$ is a discrete random variable with possible values $\{x_1, x_2, \ldots, x_n\}$ and given a constraint function $b(x_i)$ and associated constraint set $C$, then*

$$p(x_i) \leq U(x_i) \stackrel{\text{def}}{=} 1 - \sum_{j \neq i} b(x_j) \tag{1}$$

*where $U(x_i)$ is a sharp upper bound for $p(x_i)$ for all $i$ for all $p \in C$.*

**Proof:** Since probabilities sum to one we have

$$p(x_i) = 1 - \sum_{j \neq i} p(x_j) \leq 1 - \sum_{j \neq i} b(x_j) = U(x_i),$$

so $U(x_i)$ is an upper bound. For an arbitrary $x_i$ define $p^*(x)$ (depending on $x_i$) to be a probability distribution:

$$p^*(z) = \begin{cases} U(x_i) & z = x_i \\ b(z) & z \neq x_i \end{cases}$$

Thus $p*(z)$ is a probability distribution over the values for $x$ which achieves its upper bound at $z = x_i$ and satisfies all lower bound constraints. $\square$

---

[1] We use $\int$ as a general summation symbol for both continuous and discrete variables as appropriate.

Note that when the lower bounds for a variable sum to one, the lower bound equals the upper bound, and the interval for a probability value collapses to a single value. The definitions and theorem are analogous for the conditional case.

The next two theorems provide the fundamental mechanisms for calculating bounds for new probability distributions based on bounds on the input probabilities in the form described above.

**Theorem 2 (Marginalization)** *Given lower bound constraint functions $b(x|y)$ and $b(y)$ for all values of $x$ and $y$ and associated constraint sets $C$ and $D$,*

$$\forall x, b(x) = b(x|y_s)U(y_s) + \sum_{y \neq y_s} b(x|y)b(y) \qquad (2)$$

*(where $y_s$ is such that $b(x|y_s) \leq b(x|y)$ for all $y$) is a sharp lower bound for $p(x) \in C \cap D$.*

**Theorem 3 (Bayes)** *Given lower bound constraint functions $b(x|y)$ and $b(y)$ for all values of $x$ and $y$ and associated constraint sets $C$ and $D$,*

$$\forall x, y, b(y|x) = \frac{b(x|y)b(y)}{b(x|y)b(y) + U(x|y_s)U(y_s) + \sum_{y_i \neq y_s, y} U(x|y_i)b(y_i)} \qquad (3)$$

*(where $y_s$ is such that $U(x|y_s) \geq U(x|y_i)$ for all $y_i \neq y$) is a sharp lower bound for $p(y|x) \in C \cap D$.*

The proofs of these theorems is in [3]. In the following section we show how these theorems are used in influence diagrams to perform corresponding influence diagram transformations.

## 3 Interval Influence Diagrams

An influence diagram is an acyclic directed graph $D = \langle N, A \rangle$ of nodes $N$ and arcs $A$ [7]. Associated with each node $X \in N$ is a set of possible states $S_X = \{x_1, \ldots, x_n\}$. We will use the lower case $x$ to indicate one of the possible values for a node. The predecessors of a node $X$, written $\Pi_X$, are those nodes with arcs directly to $X$. Associated with each node is a conditional probability distribution over the possible states of the node, given of the possible states of its predecessors, written $p(x|s_{\Pi_X})$. In this expression $s_{\Pi_X}$ is a member of the set of combined possible outcomes for the predecessors. The set of nodes, their outcomes, arcs, and conditional probability distributions completely define a probabilistic influence diagram or belief net.

Interval influence diagrams differ from the standard influence diagram formalism in that we specify lower bounds (as in Definition 1) on the conditional probability distributions associated with each node in the network.

$$\forall X \in N, b(x|s_{\Pi_X}) \leq p(x|s_{\Pi_X}) \qquad (4)$$

Lower bounds for the probability of each possible value of the node given its predecessors is defined for all nodes in the graph. The upper bound on each probability is implicit in the lower bounds (see Equation 1).

A point-valued influence diagram encodes a *unique* joint probability distribution $P^*$ over the combined states of the nodes in the network obtained by multiplication of the conditionals in each node. If $N = \{X^1, X^2, X^3 \ldots\}$, then

$$P^*(x^1, x^2, x^3 \ldots) = \prod_i p(x^i|s_{\Pi_{X^i}}) \qquad (5)$$

where the composition of the terms on the right hand side reflect the conditional dependencies and independencies implied by the arcs in the diagram. In general, given an influence diagram D, there is a set of joint probability distributions with conditional independencies satisfying the structure of Equation 5, without regard for the component conditional and marginal probabilities at each node. We characterize these distributions with the following definition:

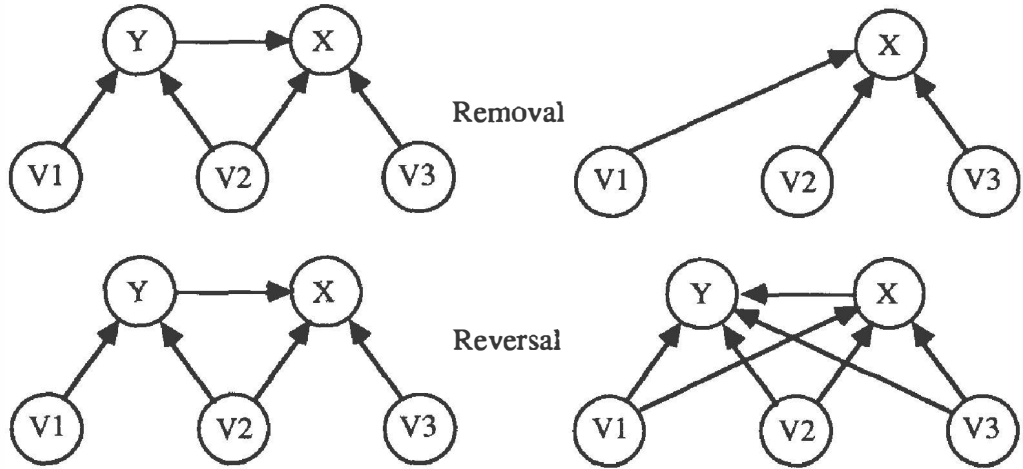

Figure 1: Graphical operations of removal and reversal

**Definition 3 (D-compatible)** *A joint probability distribution is **D-compatible** with an influence diagram $D$ of $n$ nodes when there is a labeling of nodes $X^i$ with variables $x^i$ such that Equation 5 holds.[2]*

In interval influence diagrams, the constraints on each node's probabilities further restricts the class of D-compatible joint probability distributions. In the next section we will use this definition to characterize transformations on the bounds in an influence diagram in terms of the set of joint probability distributions which are admitted by all constraints; those imposed by the graph's topology as well as those associated with lower bounds.

## 4 Transformations

The fundamental transformations to an influence diagram were defined in [13,17]. Node removal in an influence diagram corresponds to conditional marginalization, arc reversal corresponds to an application of Bayes' rule. In this section we define these operations for interval influence diagrams.

Let $X$ and $Y$ be nodes in an interval influence diagram. Let $V_1$ represent the set of predecessors of $Y$ which are not predecessors of $X$, $V_2$ represent the common predecessors of $X$ and $Y$, and $V_3$ represent the predecessors of $X$ which are not predecessors of $Y$. The following two transformations define the operations of node removal and arc reversal in interval influence diagrams, illustrated in Figure 1.

**Transformation 1 (Node Removal)** *If a node $Y$ in an interval influence diagram has a single successor $X$, $Y$ can be removed from the diagram by adding arcs from the predecessors of $Y$ to $X$ and recalculating the lower bounds on the probability distributions for $X$ as:*

$$b(x|s_{V_1}, s_{V_2}, s_{V_3}) = b(x|y_s, s_{V_2}, s_{V_3})U(y_s|s_{V_1}, s_{V_2}) + \sum_{y \neq y_s} b(x|y, s_{V_2}, s_{V_3})b(y|s_{V_1}, s_{V_2}) \qquad (6)$$

*where $y_s$ is such that $b(x|y_s, s_{V_2}, s_{V_3}) \leq b(x|y, s_{V_2}, s_{V_3})$ for all $y$.*

---

[2]A probability distribution, $P$, is D-compatible with a diagram $D$ if and only if $D$ is an *I-map* of $P$ ([15], pp. 119)

**Transformation 2 (Arc Reversal)** *If a node $Y$ in an interval influence diagram has an arc to $X$ and there is no other directed path from $Y$ to $X$, then the arc can be reversed by adding the predecessors of $X$ to $Y$ and those of $Y$ to $X$, and by*

1. *recalculating the lower bounds on the probability distributions for $Y$ as:*

$$b(y|x, s_{V_1}, s_{V_2}, s_{V_3}) = \frac{b(x|y, s_{V_2}, s_{V_3})b(y|s_{V_1}, s_{V_2})}{C + \sum_{y_i \neq y_s, y} U(x|y_i, s_{V_2}, s_{V_3})b(y_i|s_{V_1}, s_{V_2})}$$
$$C = b(x|y, s_{V_2}, s_{V_3})b(y|s_{V_1}, s_{V_2}) + U(x|y_s, s_{V_2}, s_{V_3})U(y_s|s_{V_1}, s_{V_2}) \quad (7)$$

*where $y_s$ is such that $U(x|y_s, s_{V_2}, s_{V_3}) \geq U(x|y_i, s_{V_2}, s_{V_3})$ for all $y_i \neq y$, and*

2. *recalculating the lower bounds on the probability distributions for $X$ as in Equation 6.*

These two transformations describe a mechanism for transforming a diagram from one topology to another and updating the bounds for relevant nodes in the diagram, and rely directly on theorems 2 and 3 for their semantics. We now characterize the implications of these transformations, in terms of the joint probability distribution compatible with a diagram before and after a transformation.

Let $D$ be an interval influence diagram. Associated with $D$ is a set of constraints $C_D$ on the joint probability distributions D-compatible with $D$. The constraints $C_D$ are are said to be *diagram regular* with respect to a diagram $D$ [3]. Informally, a constraint is diagram regular with respect to a set of joint probability distributions if it can be expressed as a set of independent lower bounds on the conditional probabilities of each node in the diagram as in Equation 4. An element of a diagram regular constraint set can be constructed by picking a separate conditional distribution for each variable, each of which independently satisfies Equation 4. The lower bounds are independent in the sense that $b(x|y_1)$ play no role in defining $b(x|y_2)$, for example.

Let $H$ be a single topological transformation (either node removal or arc reversal) on $D$, to produce a new diagram $D' = H(D)$. There is a corresponding transformation $T_H(C_D)$ which produces a new constraint, $C_{D'}$ on the distributions D-compatible with $D'$. Finally, there is a mapping $\overline{H}$ corresponding to $H$ which maps each joint probability distribution $p \in C_D$ to its image $\overline{H}(p)$.[3] We define

$$\overline{H}(C_D) = \{\overline{H}(p) | p \in C_D\}$$

The effectiveness of the transformations is reflected in the relationship between $\overline{H}(C_D)$, the probability distributions admitted by the underlying operation (marginalization or Bayes), and $C_{D'}$, the distributions admitted by the new constraints. These relationships are illustrated in Figure 2. Clearly we want *monotonicity*: if $p \in C_D$ then $\overline{H}(p) \in C_{D'}$ or $\overline{H}(C_D) \subseteq C_{D'}$. Any distribution that was admitted before the transformation should be admitted after.

Ideally, we would like transformations such that

$$T_H(C_D) = \overline{H}(C_D)$$

implying that the set of distributions admitted after the transformation are exactly those obtained by performing the underlying operation on the original distributions. Unfortunately the set $\overline{H}(C_D)$ is not diagram regular in that it cannot be expressed in terms of independent lower bounds on the probabilities for individual nodes, and therefore cannot be produced by transformations which are restricted to producing diagram regular constraints. However we do have the following optimality theorem [3].

---

[3] If $H$ is removal of a node x, then $\overline{H}(p) = \int_x p$. If $H$ is arc reversal, then $\overline{H}(p) = p$.

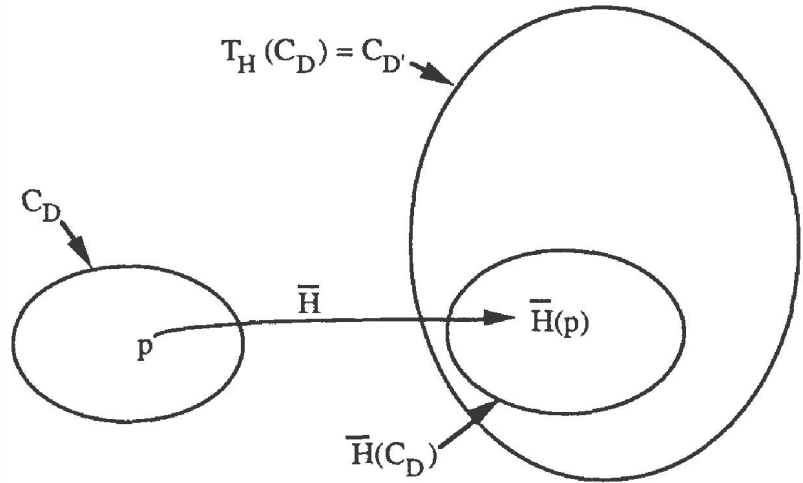

Figure 2: Mappings on probability distributions

**Theorem 4 (Minimality)** *Let $D$ be a diagram with $n$ nodes and let $C_D$ be a diagram regular constraint on the probability distributions D-compatible with $D$. Let $H$ be a single topological operation on $D$ producing $D'$, let $T_H(C_D)$ be a constraint on the distributions D-compatible with $D'$ produced by Transformation Removal or Reversal, and let $\overline{H}$ be the mapping corresponding to $H$ from distributions D-compatible with $D$ to those D-compatible with $D'$. Then*

$$T_H(C_D) = \inf_{\overline{H}(C_D) \subseteq \mathcal{C}} \mathcal{C} = \bigcap_{\overline{H}(C_D) \subseteq \mathcal{C}} \mathcal{C}$$

*where $\mathcal{C}$ is restricted to the set of diagram regular constraints with respect to $D'$.*

This theorem expresses a minimality with respect to the size of the set of admitted distributions following a transformation: it is the smallest set possible within the class of diagram regular constraints. Thus, this method will result in weaker bounds as successive transformations are applied to a diagram. In the next sections we examine this issue.

## 5 Example

Let us examine a simple example to illustrate the method. Suppose our model consists of two nodes $Y$ and $X$, where $S_Y = \{Y_1, Y_2, Y_3\}$ and $S_X = \{X_1, X_2\}$ and $\Pi X = \{Y\}$. The lower bounds database consists of the following statements:

$$\begin{array}{lll} b(y_1) = .2 & b(y_2) = .1 & b(y_3) = .4 \\ b(x_1|y_1) = .2 & b(x_1|y_2) = .2 & b(x_1|y_3) = .1 \\ b(x_2|y_1) = .0 & b(x_2|y_2) = .3 & b(x_2|y_3) = .1 \\ b(x_3|y_1) = .1 & b(x_3|y_2) = .4 & b(x_3|y_3) = .8, \end{array}$$

which imply the following intervals for each probability:

$$\begin{array}{lll} p(y_1) \in [.2, .5] & p(y_2) \in [.1, .4] & p(y_3) \in [.4, .7] \\ p(x_1|y_1) \in [.2, .9] & p(x_1|y_2) \in [.2, .3] & p(x_1|y_3) \in [.1, .1] \\ p(x_2|y_1) \in [.0, .7] & p(x_2|y_2) \in [.3, .4] & p(x_2|y_3) \in [.1, .1] \\ p(x_3|y_1) \in [.1, .8] & p(x_3|y_2) \in [.4, .5] & p(x_3|y_3) \in [.8, .8]. \end{array}$$

We now apply Equation 2 to calculate $b(x)$,

$$b(x_1) = b(x_1|y_1)b(y_1) + b(x_1|y_2)b(y_2) + b(x_1|y_3)U(y_3) = .13$$

$$b(x_2) = b(x_2|y_1)U(y_1) + b(x_2|y_2)b(y_2) + b(x_2|y_3)b(y_3) = .07$$

$$b(x_3) = b(x_3|y_1)U(y_1) + b(x_3|y_2)b(y_2) + b(x_3|y_3)b(y_3) = .41,$$

implying

$$p(x_1) \in [.13, .52]$$

$$p(x_2) \in [.07, .46]$$

$$p(x_3) \in [.41, .80].$$

Calculation of $b(x|y)$ yields the following, based on Equation 3:

$$b(y_1|x_1) = \frac{b(x_1|y_1)b(y_1)}{b(x_1|y_1)b(y_1) + U(x_1|y_2)U(y_2) + U(x_1|y_3)b(y_3)} = .2000$$

$$b(y_2|x_1) = \frac{b(x_1|y_2)b(y_2)}{U(x_1|y_1)U(y_1) + b(x_1|y_2)b(y_2) + U(x_1|y_3)b(y_3)} = .0392$$

$$b(y_3|x_1) = \frac{b(x_1|y_3)b(y_3)}{U(x_1|y_1)U(y_1) + U(x_1|y_2)b(y_2) + b(x_1|y_3)b(y_3)} = .0769,$$

implying

$$p(y_1|x_1) \in [.2000, .8839]$$

$$p(y_2|x_1) \in [.0392, .7231]$$

$$p(y_3|x_1) \in [.0769, .7608].$$

## 6 Computational Characteristics

We have implemented the system of transformations on interval influence diagrams described above in an influence diagram analysis environment [19]. It is interesting to note that the storage requirements for interval diagrams are the same as for point-valued diagrams since upper bounds are implicit in the lower bounds and need not be stored explicitly. In addition, one can see from Equation 6 that the number of floating point operations exactly the same as for removal operation in point-valued influence diagrams. For reversal (see Equation 7) there is a penalty associated with having to calculate a different denominator for each combination of $x$ and $y$ in the numerator, as well as having to calculate lower bounds on the new distribution for $x$. This increases the cost of a reversal by a factor approximately equal to the number of outcomes in $y$ over a point-valued diagram.

Though the computational costs of using the bounds approach are not severe, the more significant issue is the degradation of the bounds as transformations are performed. We can characterize the width of a bound in terms of its range—the difference between the upper and lower bounds of probabilities.

**Definition 4 (Range)** *If $x$ is a discrete random variable with possible values $\{x_1, x_2, \ldots, x_n\}$ and given $b(x_i)$ for all $i$, then*

$$R(x) \stackrel{\text{def}}{=} 1 - \sum_j b(x_j) \tag{8}$$

*is the* **range** *of $x$.*

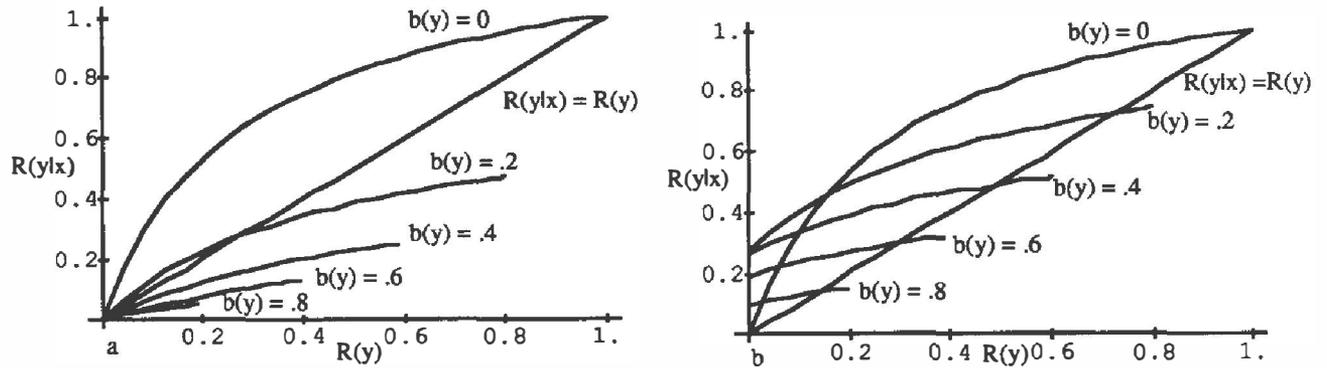

Figure 3: The range of $y$ given $x$ following reversal as a function of the range of $y$. Plot (a.) is the case with exact conditionals and (b.) is the case with bounded conditionals for $x$ given $y$.

We see from the definition that the range is independent of the particular value of the random variable. For all $i$, we have:

$$\begin{aligned} U(x_i) - b(x_i) &= 1 - \sum_{j \neq i} b(x_j) - b(x_i) \\ &= 1 - \sum_j b(x_j) \\ &= R(x) \end{aligned}$$

General equations for the behavior of ranges with the application of transformations are extremely complex. We have plotted the behavior of the ranges for some variables for marginalization (Theorem 2) and Bayes rule (Theorem 3) where both $x$ and $y$ are binary in order to provide some insight into the process.

Figure 3 shows the range of $y$ given $x$ following the reversal of $y$ using both exact and bounded conditionals as a function of the original range in $y$. Each separate curve in each plot represents a different initial lower bound for $y$. We see that the total possible range for $y$ is limited by $b(y)$. The curvature of each plot is a function of the nature of the conditionals used in the calculation, but it is apparent that in many of the cases examined with exact conditionals, the output range $R(y|x)$ can be less than the input $R(x)$. When the conditionals are not exact, then the ouput range goes up, as expected. This is represented by the shifting upward of the curves in Figure 3-b. We see in this case a substantially greater proportion of the plots are above the unit line.

Figures 4 shows the range of $x$ following the removal of $y$ using both exact and bounded conditionals as a function of the original range in $y$. As in the previous figure, each separate curve in each plot represents a different lower bound for $y$. With exact conditionals, the output range, $R(x)$ is simply a linear function of the input range $R(y)$ with a slope and intercept independent of the initial $b(y)$ chosen, so there is a single line in Figure 4-a. When conditionals are inexact we see that the curves are shifted up.

## 7 Conclusions

We have developed and implemented a system for probabilistic reasoning where the input marginal and conditional probabilities are expressed as lower bounds. The approach is efficient in terms of storage and computation, and the operations performed are optimal with respect to constraints expressed as lower bounds.

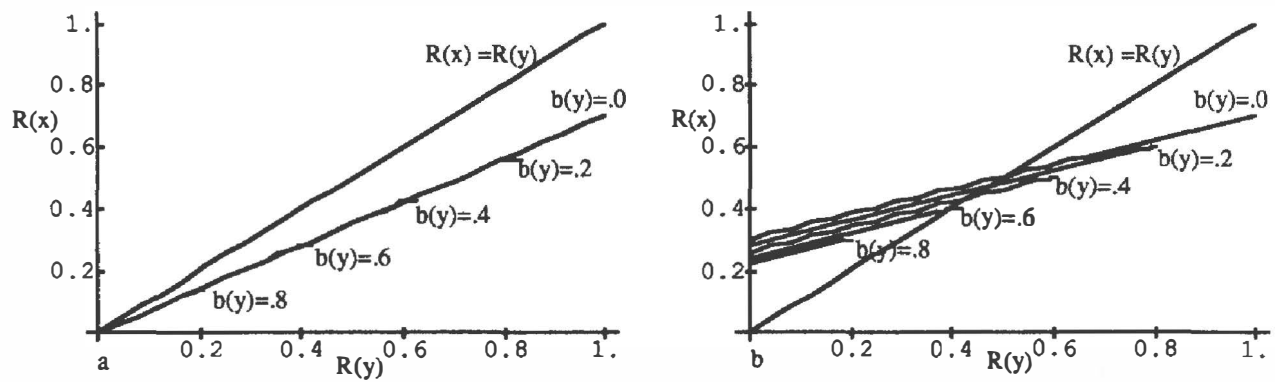

Figure 4: The range of $x$ following a removal as a function of the range of $y$. Plot (a.) is the case with exact conditionals and (b.) is the case with bounded conditionals for $x$ given $y$.

In future work we will attempt to characterize more precisely the degradation in conclusions imposed by use of lower bounds instead of generalized constraints. In addition, based on assessments of the method for probabilistic inference, we will be extending the method to consider sets of admissible decisions given bounds on distributions over expected utilities. This topic, with notable variations, has been studied by a long line of AI researchers and decision theorists [10,1,16,11]. It is hoped our computational approach will shed light and make operable these theories of decison making under incomplete information.